\documentclass[conference]{IEEEtran}
\IEEEoverridecommandlockouts
\usepackage{cite}
\usepackage{amsmath,amssymb,amsfonts}
\usepackage{algorithmic}
\usepackage{graphicx}
\usepackage{color,soul}
\usepackage{ulem}
\usepackage{array}
\usepackage{textcomp}
\usepackage{xcolor}
\usepackage{multirow}
\def\BibTeX{{\rm B\kern-.05em{\sc i\kern-.025em b}\kern-.08em
    T\kern-.1667em\lower.7ex\hbox{E}\kern-.125emX}}
\begin{document}

\title{MGSA: Multi-Granularity Graph Structure Attention for Knowledge Graph-to-Text Generation\\
}

\author{
% \IEEEauthorblockN{1\textsuperscript{st} Shanshan Wang}
\IEEEauthorblockN{Shanshan Wang}
\IEEEauthorblockA{\textit{Beijing Jiaotong University} \\
Beijing, China \\
WShan0207@163.com}
\and
% \IEEEauthorblockN{2\textsuperscript{nd} Chun zhang}
\IEEEauthorblockN{Chun Zhang}
\IEEEauthorblockA{\textit{Beijing Jiaotong University} \\
Beijing, China \\
chzhang1@bjtu.edu.cn}
\and
% \IEEEauthorblockN{3\textsuperscript{rd} Ning zhang}
\IEEEauthorblockN{Ning Zhang}
\IEEEauthorblockA{\textit{Beijing Jiaotong University} \\
Beijing, China \\
nzhang1@bjtu.edu.cn}
}

\maketitle

\begin{abstract}
The Knowledge Graph-to-Text Generation task aims to convert structured knowledge graphs into coherent and human-readable natural language text, a crucial step in making complex data accessible to non-expert users. Recent efforts in this field have focused on enhancing pre-trained language models (PLMs) by incorporating graph structure information to capture the intricate structure details of knowledge graphs. However, most of these approaches tend to capture only single-granularity structure information, concentrating either on the relationships between entities within the original graph or on the relationships between words within the same entity or across different entities. This narrow focus results in a significant limitation: models that concentrate solely on entity-level structure fail to capture the nuanced semantic relationships between words, while those that focus only on word-level structure overlook the broader relationships between original entire entities. To overcome these limitations, this paper introduces the multi-granularity graph structure attention (MGSA), which is based on PLMs. The encoder of the model architecture features an entity-level structure encoding module, a word-level structure encoding module, and an aggregation module that synthesizes information from both structure. This multi-granularity structure encoding approach allows the model to simultaneously capture both entity-level and word-level structure information, providing a more comprehensive understanding of the knowledge graph's structure information, thereby significantly improving the quality of the generated text. We conducted extensive evaluations of the MGSA model using two widely recognized KG-to-Text Generation benchmark datasets, WebNLG and EventNarrative, where it consistently outperformed models that rely solely on single-granularity structure information, demonstrating the effectiveness of our approach.
\end{abstract}


\section{Introduction}
Knowledge Graphs (KGs) \cite{b1, b2, b3} are structured data storage formats used to represent knowledge and information, with strong capabilities in data integration and information retrieval. Their graph-structured representation significantly enhances the reasoning capabilities between pieces of knowledge. However, the graph-structured nature of KGs differs greatly from natural language text, making it more difficult for humans to directly comprehend the information they contain. 
\begin{figure}[htbp]
\centerline{\includegraphics[width=0.48\textwidth]{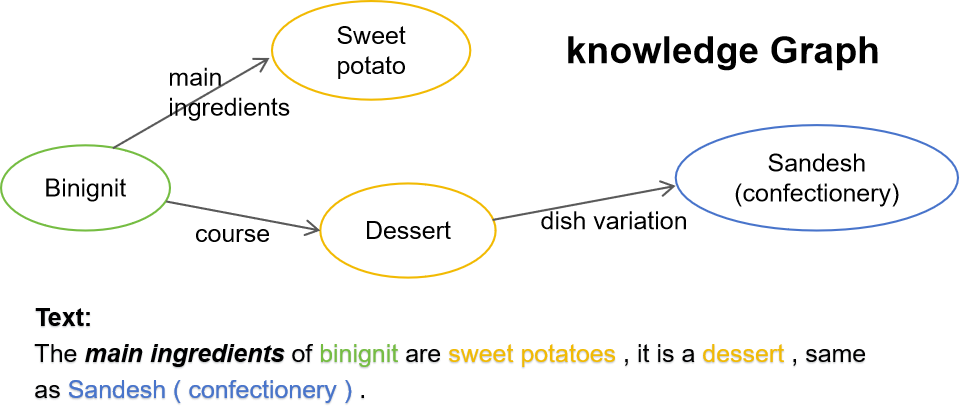}}
\caption{The knowledge graph and its corresponding description text, with italic bold indicating relationship labels and colored highlights marking entities.}
\label{fig:1}
\end{figure}
Therefore, the task of KG-to-text generation aims to transform structured KGs into human-readable natural language text, serving as a bridge between KGs and natural language. As illustrated in Fig. \ref{fig:1}, a given KG and its corresponding natural language description are presented. The task of KG-to-text  has a wide range of applications, such as question-answering systems \cite{b4, b5}, dialogue generation or dialogue agents \cite{b6}, and event narration \cite{b7}, among others.

KG-to-text generation typically requires encoding the KG so that the model can understand the information it contains, thereby generating text that accurately describes the KG. Unlike AMR-to-Text Generation \cite{b8}, which involves a more restrictive space where the graph follows predefined dense connection templates, the sparsity of KGs makes it difficult for typical text generation models to align the relationships between the KG and the target generated text. Given that PLMs have already learned rich linguistic knowledge, contextual information, and semantic relationships from large-scale pre-training corpora, recent works on KG-to-text generation \cite{b9,b10,b11} have achieved state-of-the-art (SOTA) results by leveraging PLMs. These models linearize the KG into a token sequence as input to the model, transforming the KG-to-text task into a sequence-to-sequence (seq2seq) task by fine-tuning PLMs or adding additional pre-training tasks.

Although fine-tuning PLMs has yielded encouraging results, several issues remain. First, linearizing the KG as input leads to the loss of graph structure information. Some works \cite{b12,b13,b10,b14} have attempted to add additional embedding information to the linearized sequence or introduce graph-aware modules into the model's encoder to capture the graph structure. However, these methods either focus solely on entity-level or word-level structure, without simultaneously considering both granularities of structure information. Second, when linearizing the KG, the order of the triples is randomly arranged based on the original dataset, without considering human language conventions. Our proposed model not only integrates both granularities of structure information but also clusters and arranges the linearized KG according to human language habits.

Our proposed MGSA model consists of three key modules: (1) Entity-level structure encoding module: A relative position matrix and an adjacency matrix are designed to capture entity-level structure attention information. (2) Word-level structure encoding module: A word-level relative position matrix is designed to capture word-level structure information both within and between entities. (3) Aggregation module: The aggregation module integrates the outputs of the two granularity encoding modules, and its output is used as part of the input to the decoder for generating the target text.

We evaluated the MGSA model on two popular KG-to-text generation datasets, WebNLG \cite{b15} and EventNarrative \cite{b7}, and it demonstrated better overall performance than the baseline models. Our main contributions are as follows:

\begin{itemize} \item We propose a novel KG-to-text generation model based on multi-granularity graph structure attention. By integrating graph structure information at multiple granularities, the model fully understands the information contained in the KG, thereby generating higher-quality text.

\item Based on our model, we further evaluated the effectiveness of the two granularity-level modules, providing additional evidence of the effectiveness of integrating multi-granularity structure information for encoding KG structure.

\item Through case analysis, we explored the factors influencing the model's performance. In addition to intrinsic factors within the model, we discovered that external factors, such as the quality of the dataset, also impact the model's effectiveness.

\item Experiments conducted on two datasets achieved more competitive results compared to the baseline models.

\end{itemize}

\section{Related Work}
\subsection{KG-to-Text Generation} 
KG-to-text generation is a subset of Data-to-Text generation \cite{b16} and a branch of Natural Language Generation (NLG) \cite{b17}. Unlike traditional NLG tasks (seq2seq), where the input is a linear sequence, a KG is structured as a graph rather than a linear sequence. Therefore, models designed for conventional NLG tasks cannot be directly applied to KG-to-text generation. To address this challenge, recent work has primarily focused on two directions: one direction utilizes Graph Neural Networks (GNNs) \cite{b18,b19,b20,b21,b22}, which encode node information in the graph through neighborhood aggregation and then decode the encoded results to generate the corresponding textual description. Alternatively, some approaches directly modify the encoder part of Transformer models \cite{b23}, combining them with Graph Attention Networks (GATs) \cite{b24}, enabling the model to directly compute self-attention information for all nodes in the graph. The other direction leverages the success of PLMs in NLG tasks, such as BART \cite{b25}, T5 \cite{b26}, and GPT \cite{b27}. These methods fine-tune PLMs by linearizing the structured KG into a token sequence, using the linearized KG nodes as input to generate sentences, thus transforming the graph-to-text task into a seq2seq task. However, the drawback of this approach is that the original graph structure information is lost during the linearization process.

Although explicitly encoding graph structure information with GNNs has been shown to be effective, methods based on PLMs have demonstrated superior results compared to GNNs. Therefore, we also chose to explore the PLM-based approach for the KG-to-text generation task.

\subsection{Structure Information for PLMs} 
Recent work has focused on injecting the structure information of KGs into PLMs. Studies such as \cite{b28} and \cite{b29} incorporate additional embeddings into the linearized sequence, primarily to capture token-level structure information in the KG. In \cite{b12} and \cite{b11}, a relative distance matrix is designed based on the relative distances between words or entities, and this matrix is integrated into the attention mechanism of the encoder to capture word-level or entity-level structure information from the KG. Meanwhile, \cite{b13,b10,b30} design a graph structure-aware module within the encoder to capture entity-level structure information from the KG. Additionally, \cite{b31} and \cite{b32} combine GNNs with PLMs to align the entity-level graph structure of KGs with the token-level semantic information in the linear sequence.

Overall, the aforementioned work either incorporates word-level structure information, entity-level structure information, or directly integrates entity-level structure information with the semantic information of linearized sequences. However, these approaches do not simultaneously consider the structure information between entities and the structure information between words within the same entity or across different entities in the KG. Inspired by \cite{b22} and \cite{b33}, which explore multi-granularity structure information in KGs using GNNs, we also consider both entity-level and word-level structure information when incorporating graph structure into PLMs. Experimental results demonstrate the effectiveness of our model.

\begin{figure*}
\centering
{\includegraphics[width=0.9\textwidth]{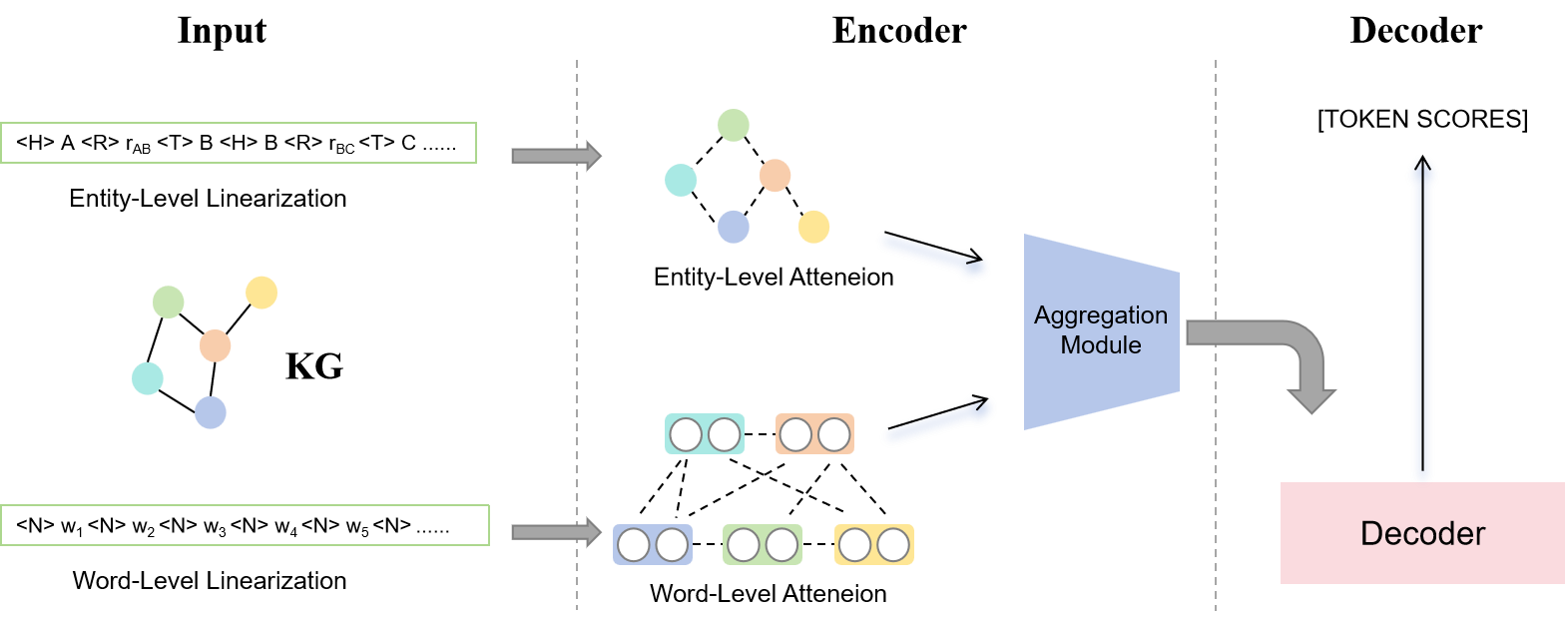}}  
\caption{Overall model architecture. The input consists of two levels of granularity in the linearization process. The encoder module incorporates structure attention at both granularities, while the decoder module follows the standard Transformer decoder structure.}
\label{fig:2}
\end{figure*}

\section{Methodology}
\subsection{Problem Definition and Model Architecture}
KG-to-text generation aims to convert a set of triples into natural language text. For a given KG \( \mathcal{G} = \{(h_i, r_{ij}, t_j) \mid h_i, t_j \in V, r_{ij} \in R\} \), where V is the set of all nodes in the graph and R is the set of all relation labels in the graph, the triples are linearized into a token sequence \( \mathcal{G}_{\text{linear}} = (x_1, x_2, \dots, x_n) \) where \(x_i\) represents the i-th token in the linear sequence and n represents the number of tokens in the sequence. The linearized sequence is input into the model to generate the corresponding text sequence \( T = (t_1, t_2, \dots, t_k) \), where \(t_i\) represents the i-th token generated by the model and \(k\) represents the length of the text.

The overall architecture of the model follows an encoder-decoder structure based on BART. As illustrated in Fig. \ref{fig:2}, the encoder section is designed to fully capture the information embedded in the graph by utilizing two granularity-specific encoding modules: a word-level module and an entity-level module, along with an aggregation module. The entity-level module captures entity-level structure information using a relative position matrix between nodes in the graph, while the word-level module captures semantic information between words within the same node or across different nodes using a relative position matrix between words. The aggregation module is responsible for fusing the encoding vectors from both granularities. In the decoder section, the model adopts a standard Transformer \cite{b34} decoder architecture, where the output from the encoder serves as part of the hidden input to the decoder for generating the target text.

\subsection{Entity-level Structure Encoding Module}
This module encompasses the linearization of entity-level structure in the KG, the design of a relative position matrix, and the incorporation of this relative position matrix into the self-attention weight calculations to effectively capture the structure information.
\subsubsection{Entity-level Linearization}
For a given KG in the form of a set of triples, the linearization process follows the method in existing work \cite{b13}. All triples are concatenated together, with special tokens \(\texttt{<H>}\), \(\texttt{<R>}\), and \(\texttt{<T>}\) used to denote the head, relation, and tail of each triple, respectively.
\subsubsection{Linear Attention}
The linearized sequence of the KG is then fed into the module, where it first undergoes self-attention computation. This process captures the global information of the linear sequence by calculating the global attention across all tokens in the sequence. Let the linearized sequence be denoted as $X^E$, and the computation process of the linear self-attention is as follows:
\begin{equation}
X_{lin} = \sigma{\left(\frac{QK^\top}{\sqrt{d}}\right)}V,
\end{equation}
where \( Q = X^EW^Q \), \( K = X^EW^K \), \( V = X^EW^V \), and \( W^Q \), \( W^K \), \( W^V \) are learnable weight parameters. \( d \) is the dimensionality of the vectors. $\sigma(\cdot)$ denotes the softmax function.
\subsubsection{Entity-level Graph Structure Attention}
While linear attention primarily captures the global information of linear sequences, it is insufficient to obtain the structure information of entities and relations in the KG. To further capture the structure information between entities and relations, the vectors $X_{lin} \in \mathbb{R}^{n \times d}$ obtained from linear attention are first transformed into entity and relation vectors $X_p \in \mathbb{R}^{m \times d}$ through a pooling layer.
\begin{equation}
X_p = \text{pooling}(X_{lin})
\end{equation}
Here, $n$ and $m$ represent the number of tokens in the linear sequence and the number of entities and relations in the KG, respectively. The pooling operation employs mean pooling.

Transforming the KG into a bipartite graph, as illustrated in Fig. \ref{fig:3}a. The nodes on the left side represent all entities from the original KG, while the nodes on the right side represent the relations. The relative position matrix $R^E$ is generated based on the connection types between nodes in the bipartite graph:
\begin{equation}
R^E_{ij} = \begin{cases}
1, & \text{if i and j are neighboring entities,} \\
2, & \text{if (i, j) is an (entity, edge) pair,} \\
3, & \text{if (i, j) is an (edge, entity) pair,}
\end{cases}
\end{equation}

The connection types are defined based on the neighboring entities or relations in the original KG. By assigning different weights to different connection types, the relative position matrix $R^E$ is derived. Using this relative position matrix $R^E$, the graph structure attention between entities is calculated:
\begin{figure}[htbp]
\centering
{\includegraphics[width=0.5\textwidth]{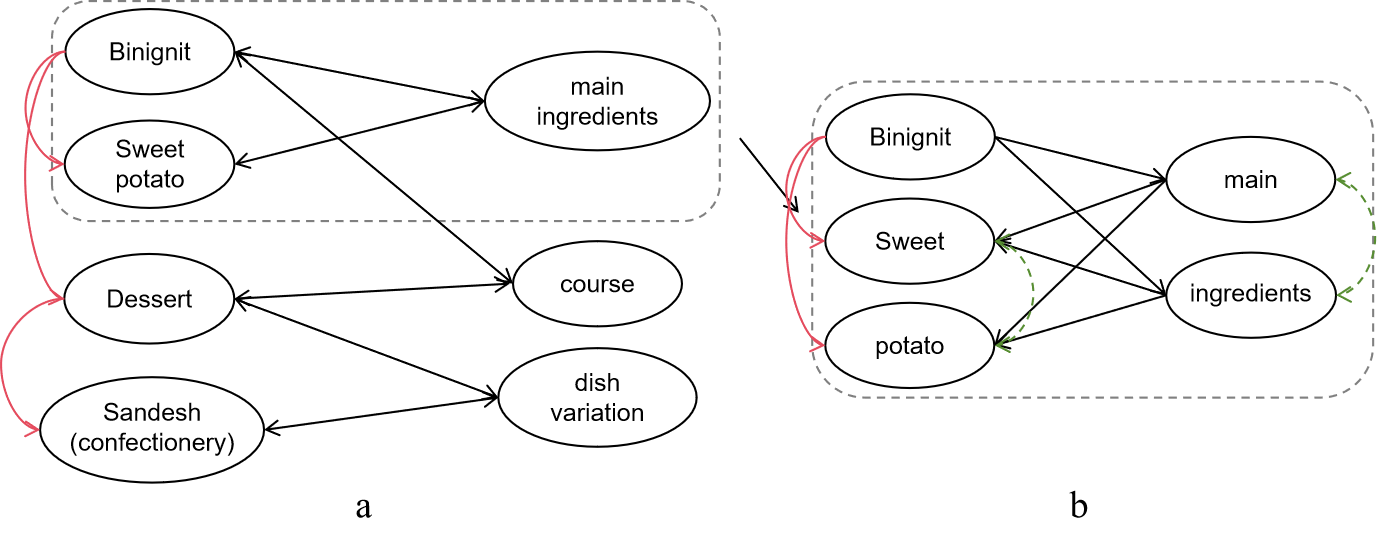}}  
\caption{Entity-level linearization and word-level linearization.}
\label{fig:3}
\end{figure}
\begin{equation}
X_g = \sigma{\left(\frac{Q^p{K^p}^\top}{\sqrt{d}} + A + \gamma{\left(R^E\right)}\right)}V^p,
\end{equation}
Here $Q^p$, $K^p$, $V^p$ are constructed from $X_p$ by multiplying it with their corresponding learnable parameter \( W^{Q^p} \), \( W^{K^p} \), \( W^{V^p} \). $A \in \mathbb{R}^{m \times m}$ is the adjacency matrix corresponding to the bipartite graph, and $\gamma(\cdot)$ maps the relative distances between entities or relations to a vector space. Finally, the entity-level structure attention representation is combined with the linear attention representation through a residual connection, producing the output of the entity-level structure encoding module.
\begin{equation}
\widetilde{X}^E = \text{gather}(X_g) + X_{lin}
\end{equation}
Here, the function $\text{gather}(\cdot)$ remaps the $m$-dimensional node representations back to the $n$-dimensional token representation space.
\subsection{Word-level Structure Encoding Module}
\subsubsection{Word-level Linearization}
An entity or relation is usually composed of several words. In word-level linearization, these entities or relations are split into individual words, with each word marked by a special token \texttt{[N]}. The words are then concatenated in the order of the original set of triples.
\subsubsection{Word-level Graph Structure Attention}
In the entity-level bipartite graph, each node consists of several words. Based on this, the graph is transformed into a word-level graph, as illustrated in Fig. \ref{fig:3}b, where each node from the original bipartite graph is split into multiple nodes, with each node containing only one word. When calculating word-level structure attention, the word-level graph is used to generate the word-level relative position matrix $R^N$ based on the shortest path distances between words:
\begin{equation}
R^N_{ij} = \begin{cases}
\infty, & \text{if } \delta(n_i, n_j) = \infty \text{ and } \delta(n_j, n_i) = \infty, \\
\text{encode}(p), & \text{if } (n_i, n_j) \in \text{SAME}_p, \\
\delta(n_i, n_j), & \text{if } \delta(n_i, n_j) \leq \delta(n_j, n_i), \\
-\delta(n_j, n_i), & \text{if } \delta(n_i, n_j) > \delta(n_j, n_i),
\end{cases}
\end{equation}
Here, $\delta(n_i, n_j)$ represents the shortest path distance from word node $n_i$ to word node $n_j$. If $n_i$ to $n_j$ are unreachable from each other, their shortest path distance is set to $\infty$. The term $\text{SAME}_p$ indicates words that belong to the same entity. The function $\text{encode}(\cdot)$ maps the distance between words from the same entity to a value outside the range of the $\delta(\cdot)$ function. The encoding function is defined as $\text{encode}(p) := \text{sgn}(p) \cdot \delta_{\text{max}} + p$, where $\delta_{\text{max}}$ is the diameter of the graph, representing the maximum shortest path length between two words. Let the linearized sequence be denoted as $X^N$, the word-level graph structure attention is calculated as follows:
\begin{equation}
X^W = \sigma{\left(\frac{Q^N{K^N}^\top}{\sqrt{d}} + \gamma{\left(R^N\right)}\right)}V^N,
\end{equation}
Here, $Q^N$, $K^N$, $V^N$ are constructed from $X^N$ by multiplying it with their weight matrices.

\subsection{Aggregation Module}
After encoding through the two granularity-specific module, we obtain the entity-level and word-level graph structure encoding vectors, which are then concatenated:
\begin{equation}
c = \left[ \widetilde{X}^E \, \| \, \lambda X^W \right]
\end{equation}
Here, $\lambda$ is a hyperparameter that controls the contribution of the word-level structure encoding vector. For the vector $c$, the overall attention is calculated to fuse the graph structure information at both granularities as follows:
\begin{equation}
X^c = \sigma{\left(\frac{Q^c{K^c}^\top}{\sqrt{d}}\right)}V^c,
\end{equation}
Here, $Q^c$, $K^c$, $V^c$ are constructed from $c$ by multiplying it with their weight matrices. The representation is then enhanced through a residual connection, followed by a two-layer feedforward network (FF) to adjust the dimensionality of the vector representation. Finally, the output representation $O$, which fuses the information from both granularities, is obtained through another residual connection and layer normalization (LN) as follows:
\begin{equation}
O = LN(FF(X^c + c) + X^c)
\end{equation}

\subsection{Loss Function}
The model's decoder adopts the standard Transformer decoder structure. For a given KG $\mathcal{G}$, the loss for text generation is calculated using the negative log-likelihood (NLL) as follows:
\begin{equation}
\mathcal{L} = - \sum_{i=1}^{k} \log p \left( t_i \mid t_1, \dots, t_{i-1}; \mathcal{G} \right)
\end{equation}
Here, $p$ represents the generation probability of each token.

\section{Experiments}
\subsection{Experiment Setup}
\subsubsection{Dataset}
The experiments were conducted on two datasets designed for the KG-to-text generation task: WebNLG v2.0 and EventNarrative. WebNLG consists of crowdsourced RDF data manually created by human annotators. Each example in the dataset contains up to seven triples and one or more reference texts. For comparison with previous work, version 2.0 was used in our experimental analysis. EventNarrative extracts events from EventKG\cite{b35} and enriches each event with additional data from Wikidata\cite{b3}, including related attributes and objects. Detailed textual descriptions related to these events are then obtained from Wikipedia. This process ensures a close correspondence between the KG and the text, resulting in a large-scale, high-quality parallel dataset. Due to computational resource limitations, ablation studies and further analysis experiments were only conducted on the WebNLG dataset, while the optimal model obtained from WebNLG was validated and tested on the EventNarrative dataset. Table \ref{tab:1} presents the official train/validation/test splits for both datasets.
\begin{table}[ht]
\centering
\caption{Dataset Splits for WebNLG and EventNarrative.}
\begin{tabular}{|c|c|c|c|}
\hline
\textbf{Dataset} & \textbf{Train} & \textbf{Valid} & \textbf{Test} \\ \hline
WebNLG & 34,352 & 4,316 & 4,224 \\ \hline
EventNarrative & 179,543 & 1,000 & 22,441 \\ \hline
\end{tabular}
\label{tab:1}
\end{table}
\subsubsection{Data Processing}
Before linearizing the KG, we clustered the set of triples based on their head entities. This ensures that information related to the same entity is not only adjacent in the graph structure but also in the linear sequence, thus preserving the completeness and coherence of the generated text in describing the entity. This approach aligns with human language habits—when we speak, we tend to focus on one topic at a time and move on to the next only after finishing the current one, rather than interweaving multiple topics in a disorganized manner.
\subsubsection{Parameter Settings}
In the experiments, we used the pre-trained \texttt{bart-base} checkpoint from Hugging Face to initialize the parameters of the entity-level module and word-level module. When training on the WebNLG dataset, we set the number of epochs to 40, the batch size to 16, the learning rate to $2\text{e-}5$, the number of warm-up steps to 1600, the beam width to 5, and chose Adam as the optimizer with $\epsilon$ set to $1\text{e-}8$. The maximum input length for the linear sequence was set to 256, and the maximum generation length for the text sequence was set to 128. The hyperparameter $\lambda$, which controls the word-level graph structure vectors, was set to 0.5. During evaluation, we followed existing KG-to-text work and used BLEU4\cite{b36}, METEOR\cite{b37}, and ROUGEL\cite{b38} scores as evaluation metrics to analyze the model's performance.
\subsubsection{Baseline Selection}
When analyzing the experimental results, we selected KG-to-text generation models that linearize KGs on the same dataset for a fair comparison, including KGPT\cite{b28}, JointGT\cite{b13}, Graformer\cite{b12}, GAP\cite{b10}, and UniD2T\cite{b14}. The KGPT model injects multi-level structure information into the linearized sequence using three special embeddings. The JointGT and GAP models design a structure-aware semantic aggregation module and a graph-aware attention module, respectively, to capture entity-level structure information. For the JointGT model, we chose the BART-based pre-trained model consistent with this experiment. The Graformer model designs a relative graph position matrix based on the Transformer model to capture word-level structure information in the KG. The UniD2T model, targeting various structured data-to-text generation tasks, captures word-level structure information in the input graph through a designed structure-enhancement module.

\begin{table}[ht]
\centering
\caption{Performance comparison on WebNLG.}
\begin{tabular}{|l|c|c|c|c|c|}
\hline
\textbf{Model} & \textbf{\#P} & \textbf{Pre+} & \textbf{BLEU4} & \textbf{METEOR} & \textbf{ROUGEL} \\ \hline
SOTA-NPT        & -        & No  & 61.00 & 42.00 & 71.00  \\ \hline
BART-base       & 140M     & Yes  & 64.55 & 46.51 & 75.13  \\ \hline
T5-base         & 220M     & Yes  & 64.42 & 46.58 & 74.77  \\ \hline
KGPT            & 177M     & Yes & 64.11 & 46.30 & 74.57  \\ \hline
JointGT         & 160M     & Yes & 65.92 & \textbf{47.15} & 76.10  \\ \hline
Graformer       & -        & -   & 61.15 & 43.38 & -      \\ \hline
GAP             & 153M     & No  & \underline{66.20} & 46.77 & \underline{76.36}  \\ \hline
UniD2T          & -        & Yes & 60.41 & 44.35 & -      \\ \hline
MGSA (ours)     & 167M     & No  & \textbf{66.45} & \underline{46.93} & \textbf{76.47}  \\ \hline
\end{tabular}
\label{tab:2}
\end{table}
\subsection{Main Results}
Table \ref{tab:2} and \ref{tab:3} present the experimental results of the proposed model on the WebNLG and EventNarrative datasets, along with the corresponding experimental data from baseline models. To ensure objectivity and fairness in comparison, all data are derived from the experimental results reported in the original papers. SOTA-NPT\cite{b39} denotes the state-of-the-art (SOTA) model without any pre-training. BART-base and T5-base are the unmodified base models. \textbf{\#P} indicates the number of model parameters, and \textbf{Pre+} denotes whether the model underwent additional pre-training. Bolded values represent the highest scores under the current evaluation metrics, while underlined values indicate the second-highest scores.

\begin{table}[ht]
\centering
\caption{Performance comparison on EventNarrative.}
\begin{tabular}{|l|c|c|c|}
\hline
\textbf{Model} & \textbf{BLEU4} & \textbf{METEOR} & \textbf{ROUGEL} \\ \hline
BART-base      & 31.38 & 26.68 & 62.65  \\ \hline
T5-base        & 12.80 & 22.77 & 52.06  \\ \hline
JointGT & 31.19 & 26.58 & \textbf{64.91}  \\ \hline
GAP            & \underline{35.08} & \underline{27.50} & 64.28  \\ \hline
MGSA (ours)    & \textbf{35.22} & \textbf{27.61} & \underline{64.46}  \\ \hline
\end{tabular}
\label{tab:3}
\end{table}
For the WebNLG dataset, our model outperforms the SOTA-NPT model, which has no pre-training, by 5.45\%, 4.93\%, and 5.47\% in BLEU4, METEOR, and ROUGEL metrics, respectively. Additionally, Compared to the BART-base and T5-base models, it achieves approximately 2\% and 1.5\% improvements in BLEU4 and ROUGEL metrics, respectively, demonstrating that the incorporation of graph structure attention matrices enhances the performance of the KG-to-text generation task. Compared to the KGPT model, which adds multi-level structure embeddings to the linearized sequence, our model achieves 2.34\% and 1.90\% improvements in BLEU4 and ROUGEL metrics, respectively. This indicates that our approach, which integrates multi-granularity graph structure attention, is more effective than adding multi-level embedding information. In the BLEU4 metric, our model surpasses models with single-granularity structure-aware modules such as JointGT, Graformer, GAP, and UniD2T by 0.53\%, 5.30\%, 0.25\%, and 6.04\%, respectively. This suggests that our multi-granularity graph structure attention matrices can more comprehensively capture both entity-level structure information and word-level semantic information, leading to a deeper understanding of the KG and the generation of higher-quality text. The significant improvement over Graformer may be attributed to our choice of the BART model, which enhances the KG-to-text generation task more effectively than the Transformer model. As for UniD2T, we speculate that our model's focus solely on graph-structured data-to-text generation tasks, without involving other structured data, contributes to its superior performance.

For the EventNarrative dataset, our model achieves the highest scores in BLEU4 and METEOR metrics, outperforming the second-best GAP model by 0.14\% and 0.11\%, respectively, and the second-highest score in the ROUGEL metric. Compared to the BART-base model, our model shows improvements of 3.84\%, 0.93\%, and 1.81\% across the three evaluation metrics, further confirming that the integration of multi-granularity graph structure attention can achieve excellent performance even with more complex KGs.

Overall, built on top of BART, our model integrates multi-granularity graph structure attention, effectively captures the structure information of the graph. This enables the model to extract information from the given KG at multiple granularities, resulting in the generation of higher-quality text.

\subsection{Ablation Study}
Due to the limitations of experimental resources, we conducted ablation studies only on the WebNLG dataset. Through experiments with single-granularity structure attention, we analyzed the impact of the entity-level structure encoding module and the word-level structure encoding module on the model's performance.
\begin{table}[ht]
\centering
\caption{Experimental results with(w/) or without(w/o) entity-level encoding module(E/M).}
\begin{tabular}{|l|c|c|c|c|c|}
\hline
\textbf{{E/M}} & \textbf{$A$} & \textbf{$R^E$} & \textbf{BLEU4} & \textbf{METEOR} & \textbf{ROUGEL} \\ \hline
w/o   & -  & -  & 62.31 & 44.53 & 73.11 \\ \hline
w/   & \checkmark  & ×  & 64.86 & 45.37 & 75.10 \\ \hline
w/   &  × & \checkmark  & 65.34 & 46.41 & 75.85 \\ \hline
w/   & ×  & ×  & 64.53 & 45.08 & 74.77 \\ \hline
w/ (MGSA) & \checkmark  & \checkmark  & \textbf{66.45} & \textbf{46.93} & \textbf{76.47} \\ \hline
\end{tabular}
\label{tab:4}
\end{table}

\subsubsection{Entity-level}
In analyzing the impact of entity-level structure encoding on model performance, we conducted experiments on models where the entity-level structure encoding module was removed, the relative position encoding matrix was removed, the adjacency matrix was removed, and both the relative position encoding and adjacency matrices were removed. The experimental results are shown in Table \ref{tab:4}. From the results, it can be observed that both types of encoding contribute to improving the model's performance. Furthermore, the relative position matrix has a greater impact on the model's effectiveness than the adjacency matrix, with the model using only the relative position matrix outperforming the one using only the adjacency matrix by 0.52\%.
\begin{table}[ht]
\centering
\caption{Experimental results with or without word-level encoding module(W/M).}
\begin{tabular}{|l|c|c|c|c|}
\hline
\textbf{{W/M}} & \textbf{$R^W$} & \textbf{BLEU4} & \textbf{METEOR} & \textbf{ROUGEL} \\ \hline
w/o   & -  & 65.58 & 46.38 & 75.27 \\ \hline
w/   & ×  & 65.77 & 46.59 & 75.63 \\ \hline
w/ (MGSA) & \checkmark  & \textbf{66.45} & \textbf{46.93} & \textbf{76.47} \\ \hline
\end{tabular}
\label{tab:5}
\end{table}
\subsubsection{Word-level}
As shown in Table \ref{tab:5}, the difference between removing the word-level structure encoding module and removing only the word-level relative position matrix is not significant. However, when the relative position matrix is added, the model's performance improves by nearly 1\% compared to the model without the relative position matrix. Additionally, by comparing the data in Table \ref{tab:5} with that in Table \ref{tab:4}, we find that removing the entity-level module has a greater impact on the model's performance than removing the word-level module, indicating that the entity-level module contributes more to the model's overall improvement.
\begin{figure}[htbp]
\centering
{\includegraphics[width=0.5\textwidth]{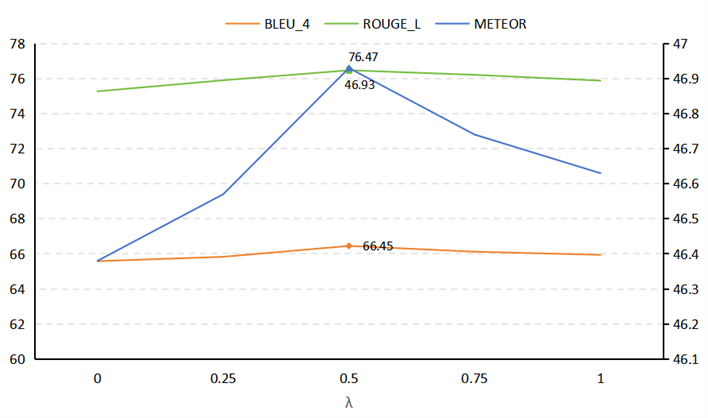}}  
\caption{Impact of $\lambda$ values.}
\label{fig:4}
\end{figure}
\subsection{Impact of $\lambda$ values}
Additionally, we explored the impact of the $\lambda$ value in the aggregation module on the model's performance. The Fig. \ref{fig:4} shows the trend of experimental results with $\lambda$ values set to 0, 0.25, 0.50, 0.75, and 1. When $\lambda = 0$, it is equivalent to removing the word-level module. From $\lambda = 0$ to $\lambda = 0.5$, the experimental results show an upward trend, reaching the maximum at $\lambda = 0.5$. From $\lambda = 0.5$ to $\lambda = 1$, the experimental results gradually decline. Although the optimal model performance may not be achieved at $\lambda = 0.5$, due to resource limitations, we selected $\lambda = 0.5$ as the optimal value in our experiments. Overall, the experimental results indicate that the value of $\lambda$ also affects the model's performance.

\subsection{Case Study}
\begin{table*}[ht]
\centering
\caption{Analysis of Generation Performance on WebNLG and EventNarrative Datasets with Different Knowledge Graph Scales.}
\begin{tabular}{|p{1.8cm}|p{6cm}|p{4cm}|p{4.7cm}|}
\hline
\textbf{Dataset} & \textbf{Knowledge Graph Triples} & \textbf{Reference Text} & \textbf{Generated Text} \\ \hline

\multirow{2}{*}{\parbox{1.8cm}{WebNLG \newline
(4 and 7 triples)}} & 
(Acharya Institute of Technology, affiliation, Visvesvaraya Technological University) \newline
(Acharya Institute of Technology, country, India) \newline
(Acharya Institute of Technology, campus, In Soldevanahalli, Acharya Dr. Sarvapalli Radhakrishnan Road, Hessarghatta Main Road, Bangalore – 560090.) \newline
(Visvesvaraya Technological University, city, Belgaum) 
 & 
The Acharya Institute of Technology 's \textbf{campus} is located in Soldevanahalli, Acharya Dr. Sarvapalli Radhakrishnan Road, Hessarghatta Main Road, Bangalore - 560090, India. It is affiliated with the Visvesvaraya Technological University in Belgaum. & 
The \hl{Acharya Institute of Technology} is located in \hl{Soldevanahalli, Acharya Dr. Sarvapalli Radhakrishnan Road, Hessarghatta Main Road, Bangalore – 560090, India}. It is \hl{affiliated} with the \hl{Visvesvaraya Technological University in Belgaum}. \\ \cline{2-4}

& 
(School of Business and Social Sciences at the Aarhus University, city, Aarhus) \newline
(School of Business and Social Sciences at the Aarhus University, affiliation, European University Association) \newline
(School of Business and Social Sciences at the Aarhus University, dean, Thomas Pallesen) \newline
(School of Business and Social Sciences at the Aarhus University, number of students, 16000) \newline
(School of Business and Social Sciences at the Aarhus University, country, Denmark) \newline
(School of Business and Social Sciences at the Aarhus University, latin name, Universitas Aarhusiensis) \newline
(School of Business and Social Sciences at the Aarhus University, established, 1928)
 & 
The School of Business and Social Sciences at the Aarhus University is located in Aarhus, Denmark and it was established in 1928. Its dean is Thomas Pallesen and it has 16,000 students. Its Latin name is ``Universitas Aarhusiensis''. It is affiliated to the European University Association. & 
The \hl{School of Business and Social Sciences at the Aarhus University in Aarhus, Denmark} was \hl{established in 1928}. It is \hl{affiliated} with the \hl{European University Association} and its dean is \hl{Thomas Pallesen}. There are 16000 students and \sout{16000 Latin speaking people}. \\ \hline

\multirow{2}{*}{\parbox{1.8cm}{EventNarrative \newline
(3 and 8 triples)}} & 
(2016 Tianjin Health Industry Park, sport, tennis) \newline
(2016 Tianjin Health Industry Park, location, Tianjin) \newline
(2016 Tianjin Health Industry Park, point in time, 2016)
 & 
The 2016 Tianjin Health Industry Park was a professional tennis tournament played on outdoor hard courts. It took place in Tianjin, China, on 2016–2016. & 
The \hl{2016 Tianjin Health Industry Park} was a professional \hl{tennis} tournament played on outdoor hard courts. It took place in \hl{Tianjin}, China, on \hl{2016 - 2016}. \\ \cline{2-4}

& 
(2019 World Para Athletics Championships, location, Dubai) \newline
(2019 World Para Athletics Championships, organizer, International Paralympic Committee) \newline
(2019 World Para Athletics Championships, sport, para athletics) \newline
(2019 World Para Athletics Championships, edition number, 9) \newline
(2019 World Para Athletics Championships, country, United Arab Emirates) \newline
(2019 World Para Athletics Championships, point in time, 2019) \newline
(2019 World Para Athletics Championships, end time, 15 November 2019) \newline
(2019 World Para Athletics Championships, follows, 2017 World Para Athletics Championships)
 & 
The 2019 World Para Athletics Championships was a para athletics track and field event organised by the World Para Athletics subcommittee of the International Paralympic Committee. It was held in Dubai, United Arab Emirates from 7 to 15 November 2019. In July 2017 during 2017 World Para Athletics Championships there were reports and speculation that London could once again hold the games in 2019 due to the success of the 2017 event and the 2012 Summer Paralympics. As of day 9, Friday 15 November 2019. & 
The 9th \hl{2019 World Para Athletics Championships}, the ninth edition of the \hl{para athletics} World Championships, was held in \hl{Dubai, United Arab Emirates} \sout{from 9 to 15 November 2019}. Dubai was selected as the host city by the \sout{International Paralympic Committee (icc)} \hl{on 9 August 2016}. The 2019 championships was the first time that Dubai hosted the World Championships. \sout{The athletics federation of africa announced dubai as the venue for the event}, and Dubai became the first city to bid for the championships in the same year. \sout{The event was scheduled to be held at the olympic} \\ \hline
\end{tabular}
\label{tab:6}
\end{table*}

To demonstrate the effectiveness of the proposed method, we analyzed the generated texts on the WebNLG and EventNarrative datasets in Table \ref{tab:6}. Two examples were provided for each dataset: one with a small-scale KGs (containing 1-4 triples) and one with a larger-scale KGs (containing more than 4 triples). As shown in the examples below, for small-scale KGs, the generated texts are almost identical to the reference texts in content and successfully convey the semantic knowledge contained in the set of triples. This explains why the earlier experimental results yielded higher scores on the WebNLG dataset, which contains more small-scale KGs. However, for larger-scale KGs, the generated texts start to exhibit issues such as missing content, hallucinations, and redundant expressions. These issues are less apparent on the WebNLG dataset. For instance, when the number of triples reaches 7, there is only one case of missing content and hallucination, with the rest of the generated content almost semantically aligned with the reference text. However, the performance on the EventNarrative dataset is significantly worse; although only one triple's content is missing, the generated text almost fails to faithfully represent the given KG, and the reference text itself also fails to fully align with the given set of triples. From this analysis, we can conclude that the model's performance is influenced not only by the scale of the KG but also by the quality of the data.

\section{Conclusion}
This paper proposes a KG-to-text generation model based on multi-granularity graph structure attention, which simultaneously considers both entity-level and word-level structure information in the knowledge graph. The model primarily consists of three modules: an entity-level structure encoding module, a word-level structure encoding module, and an aggregation module that integrates structure encoding information from both granularities. Experimental evaluations conducted on two KG-to-text benchmark datasets demonstrate that the proposed model consistently outperforms baseline models (which utilize single-granularity structure). An analysis of the factors influencing the model further explains its effectiveness in generating text from knowledge graph.


\begin{thebibliography}{00}
\bibitem{b1} Auer, Sören, et al. "Dbpedia: A nucleus for a web of open data." international semantic web conference. Berlin, Heidelberg: Springer Berlin Heidelberg, 2007.
\bibitem{b2} Bollacker, Kurt, et al. "Freebase: a collaboratively created graph database for structuring human knowledge." Proceedings of the 2008 ACM SIGMOD international conference on Management of data. 2008.
\bibitem{b3} Vrandečić, Denny, and Markus Krötzsch. "Wikidata: a free collaborative knowledgebase." Communications of the ACM 57.10 (2014): 78-85.
\bibitem{b4} Wu, Yike, et al. "Retrieve-rewrite-answer: A kg-to-text enhanced llms framework for knowledge graph question answering." arXiv preprint arXiv:2309.11206 (2023).
\bibitem{b5} Luo, Haoran, et al. "Chatkbqa: A generate-then-retrieve framework for knowledge base question answering with fine-tuned large language models." arXiv preprint arXiv:2310.08975 (2023).
\bibitem{b6} Ghanem, Hussam, Massinissa Atmani, and Christophe Cruz. "Knowledge Graph for NLG in the context of conversational agents." arXiv preprint arXiv:2307.01548 (2023).
\bibitem{b7} Colas, Anthony, et al. "EventNarrative: A large-scale event-centric dataset for knowledge graph-to-text generation." arXiv preprint arXiv:2111.00276 (2021).
\bibitem{b8} Jin, Lisa, and Daniel Gildea. "Amr-to-text generation with cache transition systems." arXiv preprint arXiv:1912.01682 (2019).
\bibitem{b9} Ribeiro, Leonardo FR, et al. "Investigating pretrained language models for graph-to-text generation." arXiv preprint arXiv:2007.08426 (2020).
\bibitem{b10} Colas, Anthony, Mehrdad Alvandipour, and Daisy Zhe Wang. "GAP: A graph-aware language model framework for knowledge graph-to-text generation." arXiv preprint arXiv:2204.06674 (2022).
\bibitem{b11} Zhao, Feng, Hongzhi Zou, and Cheng Yan. "Structure-aware Knowledge Graph-to-text Generation with Planning Selection and Similarity Distinction." Proceedings of the 2023 Conference on Empirical Methods in Natural Language Processing. 2023.
\bibitem{b12} Schmitt, Martin, et al. "Modeling graph structure via relative position for text generation from knowledge graphs." arXiv preprint arXiv:2006.09242 (2020).
\bibitem{b13} Ke, Pei, et al. "Jointgt: Graph-text joint representation learning for text generation from knowledge graphs." arXiv preprint arXiv:2106.10502 (2021).
\bibitem{b14} Li, Shujie, et al. "Unifying Structured Data as Graph for Data-to-Text Pre-Training." Transactions of the Association for Computational Linguistics 12 (2024): 210-228.
\bibitem{b15} Gardent, Claire, et al. "Creating training corpora for nlg micro-planning." 55th Annual Meeting of the Association for Computational Linguistics, ACL 2017. Association for Computational Linguistics (ACL), 2017.
\bibitem{b16} Osuji, Chinonso Cynthia, Thiago Castro Ferreira, and Brian Davis. "A Systematic Review of Data-to-Text NLG." arXiv preprint arXiv:2402.08496 (2024).
\bibitem{b17} Li, Junyi, et al. "Pre-trained language models for text generation: A survey." ACM Computing Surveys 56.9 (2024): 1-39.
\bibitem{b18} Kipf, Thomas N., and Max Welling. "Semi-supervised classification with graph convolutional networks." arXiv preprint arXiv:1609.02907 (2016).
\bibitem{b19} Guo, Zhijiang, et al. "Densely connected graph convolutional networks for graph-to-sequence learning." Transactions of the Association for Computational Linguistics 7 (2019): 297-312.
\bibitem{b20} Ribeiro, Leonardo FR, et al. "Modeling global and local node contexts for text generation from knowledge graphs." Transactions of the Association for Computational Linguistics 8 (2020): 589-604.
\bibitem{b21} Ye, Zi, et al. "A comprehensive survey of graph neural networks for knowledge graphs." IEEE Access 10 (2022): 75729-75741.
\bibitem{b22} Yang, Tianyu, Yuxiang Zhang, and Tao Jiang. "Boosting KG-to-Text Generation via Multi-granularity Graph Representations." 2022 International Joint Conference on Neural Networks (IJCNN). IEEE, 2022.
\bibitem{b23} Koncel-Kedziorski, Rik, et al. "Text generation from knowledge graphs with graph transformers." arXiv preprint arXiv:1904.02342 (2019).
\bibitem{b24} Veličković, Petar, et al. "Graph attention networks." arXiv preprint arXiv:1710.10903 (2017).
\bibitem{b25} Lewis, M. "Bart: Denoising sequence-to-sequence pre-training for natural language generation, translation, and comprehension." arXiv preprint arXiv:1910.13461 (2019).
\bibitem{b26} Raffel, Colin, et al. "Exploring the limits of transfer learning with a unified text-to-text transformer." Journal of machine learning research 21.140 (2020): 1-67.
\bibitem{b27} Radford, Alec, et al. "Language models are unsupervised multitask learners." OpenAI blog 1.8 (2019): 9.
\bibitem{b28} Chen, Wenhu, et al. "KGPT: Knowledge-grounded pre-training for data-to-text generation." arXiv preprint arXiv:2010.02307 (2020).
\bibitem{b29} Gong, Hongda, Shimin Shan, and Hongkui Wei. "Graph-to-Text Generation Combining Directed and Undirected Structural Information in Knowledge Graphs." 2023 5th International Conference on Natural Language Processing (ICNLP). IEEE, 2023.
\bibitem{b30} Wang, Tao, et al. "Improving plms for graph-to-text generation by relational orientation attention." Neural Processing Letters 55.6 (2023): 7967-7983.
\bibitem{b31} Zhao, Tianxin, et al. "Exploring the Synergy of Dual-path Encoder and Alignment Module for Better Graph-to-Text Generation." Proceedings of the 2024 Joint International Conference on Computational Linguistics, Language Resources and Evaluation (LREC-COLING 2024). 2024.
\bibitem{b32} Yuan, Shuzhou, and Michael Färber. "GraSAME: Injecting Token-Level Structural Information to Pretrained Language Models via Graph-guided Self-Attention Mechanism." arXiv preprint arXiv:2404.06911 (2024).
\bibitem{b33} Shi, Kaile, et al. "Enriched entity representation of knowledge graph for text generation." Complex \& Intelligent Systems 9.2 (2023): 2019-2030.
\bibitem{b34} Vaswani, A. "Attention is all you need." Advances in Neural Information Processing Systems (2017).
\bibitem{b35} Gottschalk, Simon, and Elena Demidova. "Eventkg: A multilingual event-centric temporal knowledge graph." The Semantic Web: 15th International Conference, ESWC 2018, Heraklion, Crete, Greece, June 3–7, 2018, Proceedings 15. Springer International Publishing, 2018.
\bibitem{b36} Papineni, Kishore, et al. "Bleu: a method for automatic evaluation of machine translation." Proceedings of the 40th annual meeting of the Association for Computational Linguistics. 2002.
\bibitem{b37} Denkowski, Michael, and Alon Lavie. "Meteor universal: Language specific translation evaluation for any target language." Proceedings of the ninth workshop on statistical machine translation. 2014.
\bibitem{b38} Lin, Chin-Yew. "Rouge: A package for automatic evaluation of summaries." Text summarization branches out. 2004.
\bibitem{b39} Shimorina, Anastasia, and Claire Gardent. "Handling rare items in data-to-text generation." Proceedings of the 11th international conference on natural language generation. 2018.
\end{thebibliography}
\end{document}